\documentclass[10pt,twocolumn,letterpaper]{article}

\usepackage{cvpr}              

\usepackage[dvipsnames]{xcolor}
\definecolor{cvprblue}{rgb}{0.21,0.49,0.74}
\usepackage[pagebackref,breaklinks,colorlinks,citecolor=cvprblue]{hyperref}

\def\paperID{2934} 
\def\confName{CVPR}
\def\confYear{2024}

\title{Dense Vision Transformer Compression with Few Samples}

\author{Hanxiao Zhang, Yifan Zhou, Guo-Hua Wang and Jianxin Wu\\
National Key Laboratory for Novel Software Technology, Nanjing University, China\\
School of Artificial Intelligence, Nanjing University, China\\
{\tt\small \{zhanghx, zhouyifan, wangguohua\}@lamda.nju.edu.cn, wujx2001@gmail.com}
}

\begin{document}
\maketitle
\begin{abstract}
Few-shot model compression aims to compress a large model into a more compact one with only a tiny training set (even without labels). Block-level pruning has recently emerged as a leading technique in achieving high accuracy and low latency in few-shot CNN compression. But, few-shot compression for Vision Transformers (ViT) remains largely unexplored, which presents a new challenge. In particular,  the issue of sparse compression exists in traditional CNN few-shot methods, which can only produce very few compressed models of different model sizes. This paper proposes a novel framework for few-shot ViT compression named DC-ViT. Instead of dropping the entire block, DC-ViT selectively eliminates the attention module while retaining and reusing portions of the MLP module. DC-ViT enables dense compression, which outputs numerous compressed models that densely populate the range of model complexity. DC-ViT outperforms state-of-the-art few-shot compression methods by a significant margin of 10 percentage points, along with lower latency in the compression of ViT and its variants.
\end{abstract}  

\section{Introduction}
\label{sec:1_intro}

Vision Transformers (ViT) have demonstrated outstanding results on various vision tasks by leveraging the Transformer structure initially proposed for natural language processing (NLP). Unlike Convolutional Neural Networks (CNN), the self-attention mechanism allows ViT to learn global representations for images without an obvious inductive bias. However, with millions to billions of parameters, ViTs can only be deployed on high-end devices, even if we only consider the inference stage. To fit large models into small devices, network compression techniques are commonly adopted to reduce computing and memory costs.
    
Although some methods have been proposed to compress ViT through either token pruning or channel pruning~\cite{rao2021dynamicvit, liang2022evit, pan2021ia, chen2021chasing, chavan2022vision, yu2022unified}, they all assume \emph{access to the entire original training set}, which is unfortunately not the case in numerous situations, especially outside of the academia. When working with datasets containing large amounts of sensitive data (\eg, medical or commercial), few-sample compression enables model pruning using few non-sensitive data and safeguards data privacy and security.

\begin{figure}
    \centering
    \includegraphics[width=0.475\textwidth]{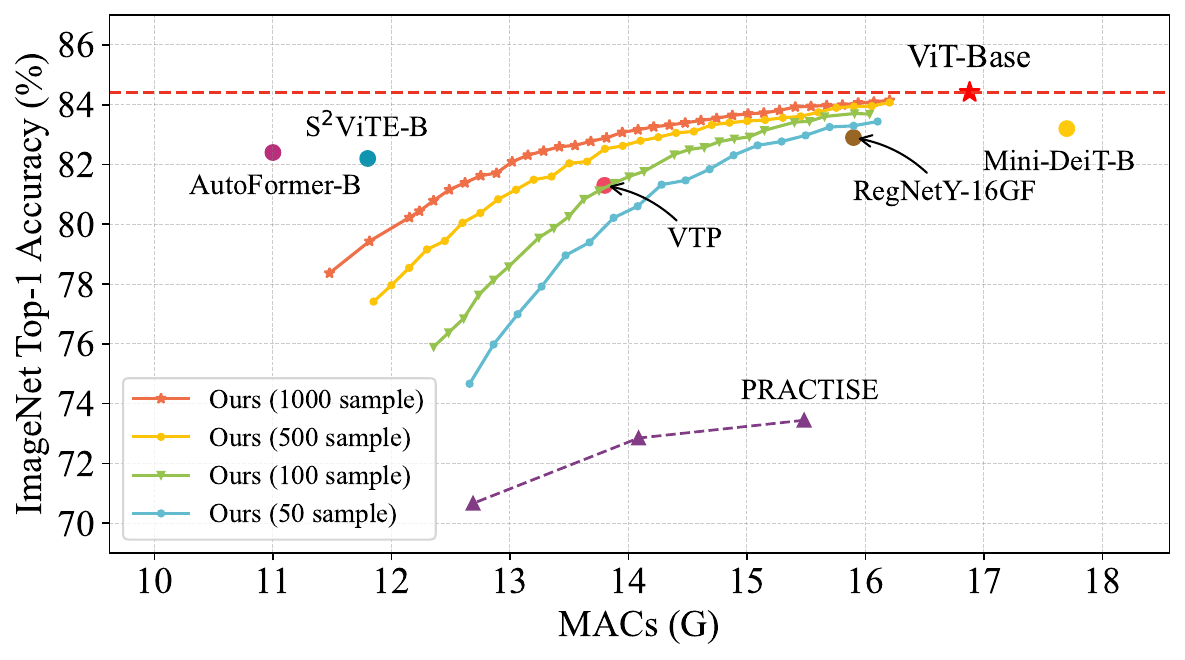}
	\caption{Top-1 accuracy vs. MACs (G) on ImageNet. Our DC-ViT was compressed with few training images (50, 100, 500, or 1000), and PRACTISE was compressed with 500 images. However, other methods were not few-shot, using 1.28 million labeled training images during compression. It is quite a success that our DC-VIT uses only \textbf{\emph{less than 0.1\%} of their training images \emph{without labels}}, but achieves slightly lower or even higher accuracy.}
	\label{fig: dense-compression}
\end{figure}

Few-shot compression has been studied for years, but existing methods focus solely on CNN compression. As Vision Transformers (ViT) is already a mainstream model, it is critical to design few-shot ViT compression methods. As shown in \cref{sec: baseline}, existing few-shot CNN compression methods do \emph{not} work well in few-shot ViT compression. The proposed DC-ViT method not only outperforms previous state-of-the-art (SOTA) methods in compressing ViTs, but also in few-shot compression of CNNs (\cf \cref{fig: dense-compression}).

Another essential demand is that the method should be both highly accurate and dense, where \emph{dense} means many models can be compressed in the target compression range, while a \emph{sparse} method can only output few models (DC-ViT vs. PRACTISE in \cref{fig: dense-compression}). If a sparse compression method can only emit two compressed models with 80\% and 110\% of the target MACs the hardware can afford, we are forced to choose the 80\% one, and 20\% of the available computing power is wasted, which will lead to lower accuracy compared to a model that can use 100\% of the compute budget. A \emph{dense} compression method will naturally overcome this drawback.
    
\begin{figure}[t]
    \centering
    \includegraphics[width=0.47\textwidth]{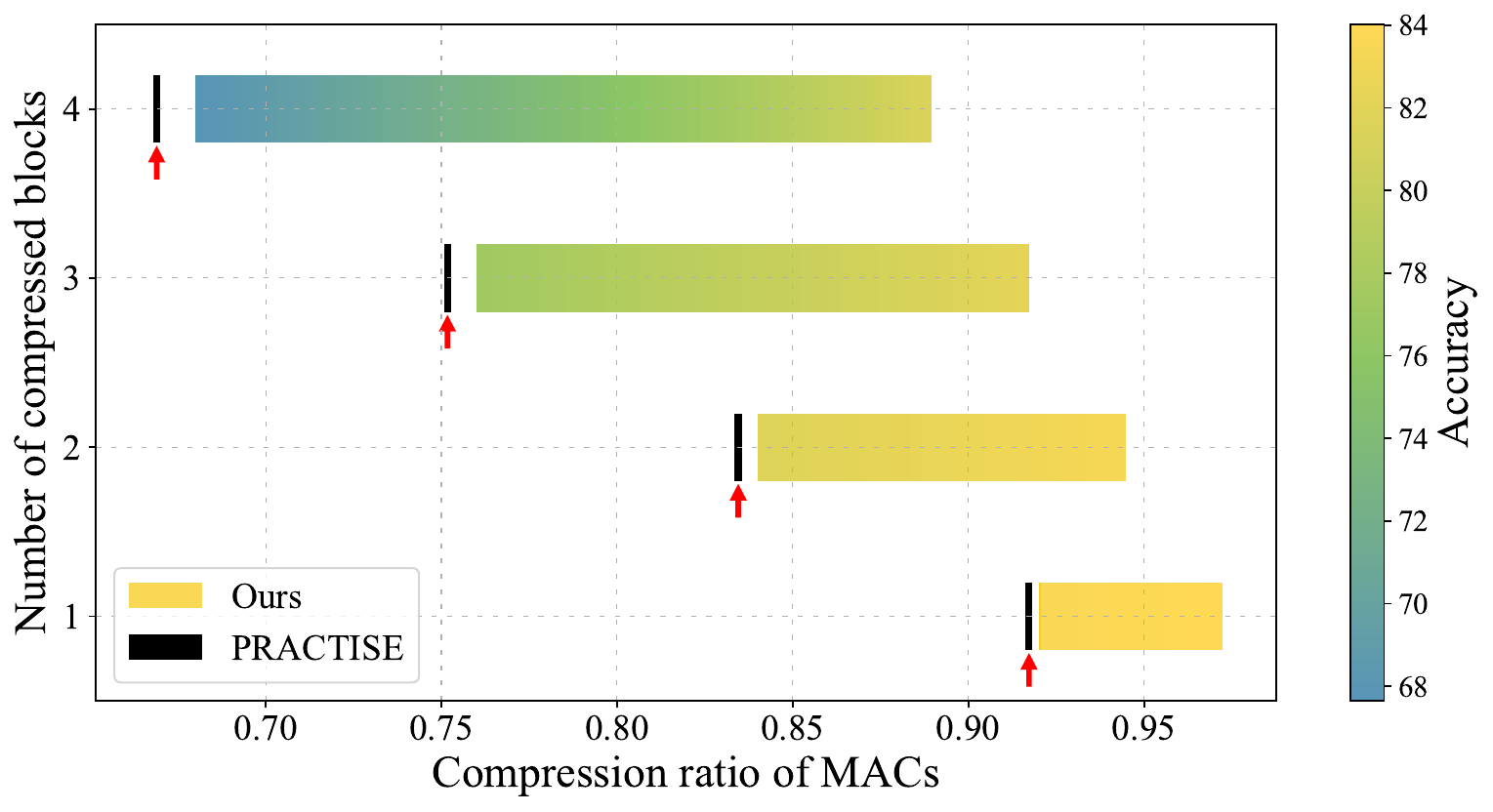}
\caption{The range of compression options attainable by our DC-ViT and PRACTISE~\citep{wang2023practical} by varying the number of compressed blocks. The black vertical lines represent the specific compression rates achieved by~\cite{wang2023practical}, highlighted by the red arrows. DC-ViT can output models densely in a wide range of compression rates for different numbers of blocks. The gradient color bar indicates the accuracy of our method at different compression rates.}
\label{fig: MACs-options}
\end{figure}

PRACTISE~\citep{wang2023practical} is a state-of-the-art few-shot CNN compression method, which outperforms other existing methods significantly in accuracy but only provides sparse compression options. As shown in \cref{fig: MACs-options}, given the target range for MACs compression, PRACTISE can only produce 4 models by dropping different numbers of blocks. While existing filter-level few-shot CNN pruning methods might attain dense compression, they exhibit substantial accuracy degradation compared to block-level pruning~\cite{wang2023practical}.

In this paper, we propose a novel and effective framework for \emph{dense} ViT compression under the few-shot setting. As shown in \cref{fig: MACs-options}, our method (denoted by the gradient color bars) can produce compressed models of various sizes that densely cover the range of MACs. We also introduce a new metric to measure the importance of a block, which matches the actual performance of the \emph{finetuned} model after compressing different blocks. To the best of our knowledge, the proposed Dense Compression of Vision Transformers (DC-ViT) is the first work in dense few-shot compression of both ViT and CNN.

As shown in \cref{fig: dense-compression}, our DC-ViT offers much \emph{denser} compression than other structured pruning methods, which means that for \emph{any} target compression ratio within a certain range, we can always find one compression setting in DC-ViT that is close enough. It significantly outperforms the SOTA few-shot compression method PRACTISE by more than \emph{10 percentage points} in top-1 accuracy.

Moreover, \cref{fig: dense-compression} further shows comparison with compressed or neural architecture searched ViT/DeiT~\cite{touvron2021training} and CNN models. Note they all used~\emph{1.28 million labeled training images} for compression, while DC-ViT used only \emph{$<0.1$\% of their training images, which are even unlabeled}. It is a surprising success that with the same MACs, DC-ViT achieves higher Top-1 accuracy than Mini-DeiT-B~\cite{MiniViT}, RegNetY-16GF~\cite{RegNetY-16GF} and VTP~\cite{VTP-KDD}, and only slightly (2--4 percentage points) lower than AutoFormer-B~\cite{AutoFormer} and $\text{S$^2$ViTE-B}$~\cite{chen2021chasing}.

Overall, our contributions are summarized as follows:
\begin{itemize}
    \item We propose DC-ViT, an effective framework that pioneers few-sample ViT compression.
    \item We advocate the benefits and necessity of dense compression in practical applications, which is achieved by dropping the attention and adjusting the MLP drop ratio.
    \item We propose generating synthetic images as a metric dataset to address the potential misalignment between the perceived recovery difficulty (indicated by the training loss) and the actual performance of finetuned models after removing different blocks. We also summarize some suggestions for few-shot ViT compression, which have shown their advantages in DC-ViT and may be useful for future work in the area.
\end{itemize}

\section{Related Works}
\label{sec:2_relatedwork}

Building upon its success in natural language processing, researchers have extended the application of Transformers to the computer vision domain, resulting in the ViT models~\cite{dosovitskiy2021an}. Many variants of ViT have been proposed for a wide range of other computer vision tasks, including object detection~\citep{carion2020end,zhu2021deformable}, semantic segmentation~\citep{zheng2021rethinking,ye2019cross}, image super-resolution~\citep{yang2020learning}, image generation~\citep{chen2021pre}, and video understanding~\citep{sun2019videobert,girdhar2019video}. However, the high computational cost limits its deployment, prompting further research into model compression techniques to address this challenge.
    
Pruning or compression has been extensively studied to reduce a network's complexity and to make it possible to migrate large networks to less powerful devices (\eg, edge devices). Previous methods of CNN compression can be broadly grouped into two categories. Unstructured pruning~\citep{dong2017learning,lee2018snip,xiao2019autoprune} typically removes unimportant weights utilizing importance metrics such as the magnitude-based metrics~\citep{hansong2016deep} or Hessian-based ones~\citep{lecun1989optimal,hassibi1992second,han2015learning}. By setting unimportant weights to zero, these methods achieve highly sparse but unstructured patterns for the network parameters. However, general-purpose GPUs do not support unstructured pruning, and their speed is in fact slow. Structured pruning solves this difficulty by completely removing some groups of parameters, such as filters~\citep{wen2016learning,li2017pruning} or blocks~\citep{lin2019towards,wang2023practical}, where the key is to evaluate the importance of different modules by certain criteria to determine which groups of parameters should be removed~\citep{wang2019dbp, wang2023practical}.
    
\begin{figure*}
    \centering
    \includegraphics[width=\linewidth]
    {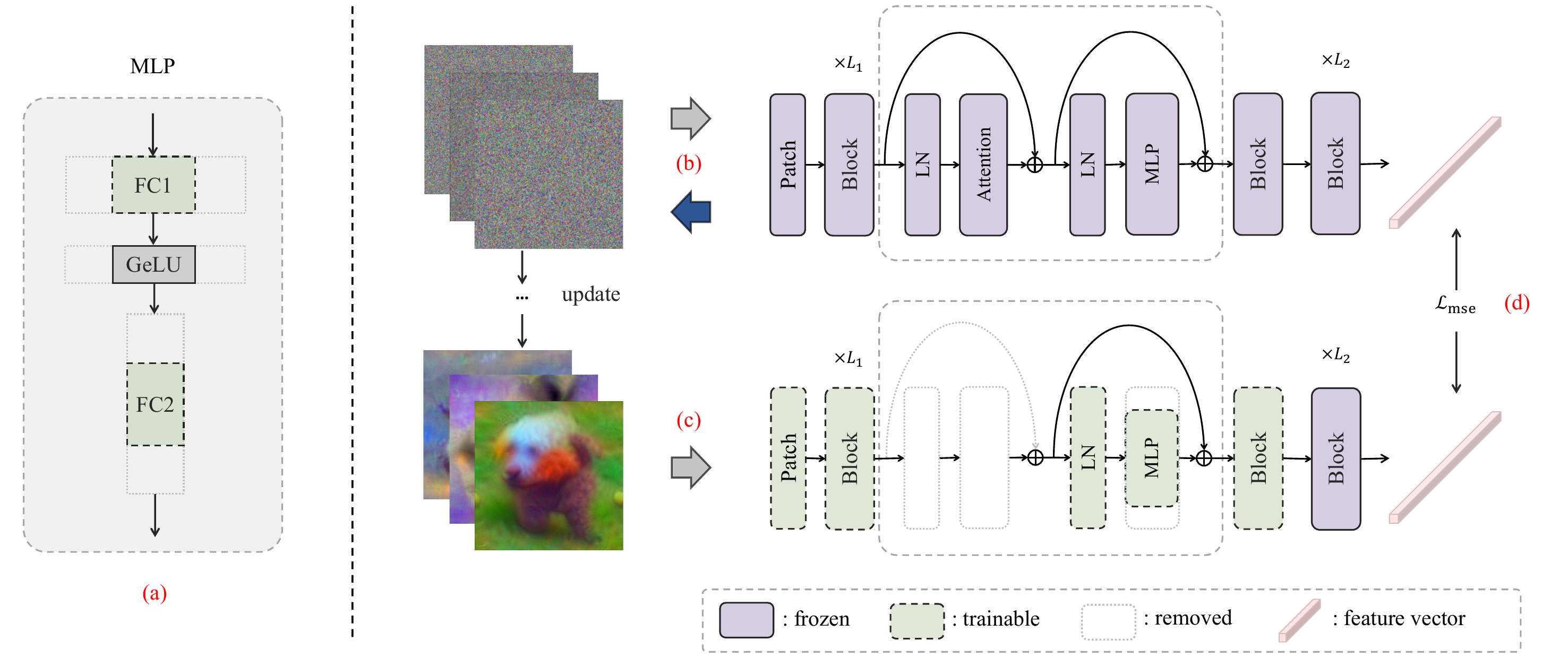}
    \caption{The DC-ViT framework. (a) Determining the network structure to achieve the target MACs reduction. (b) Generating a synthetic metric set from Gaussian noise. (c) Using the synthetic metric set to select the blocks with highest recoverability. (d) Progressive pruning and finetuning. The original model is at the top, and the pruned model is at the bottom. We drop the whole attention module but reuse part of the MLP, and use the MSE loss for feature mimicking, but \emph{only} update the front part of blocks till the next block of the compressed one.}
    \label{fig: DC-ViT framework}
\end{figure*}

As for ViT pruning, existing methods can be broadly divided into token pruning and channel pruning. Token pruning focuses on dynamically selecting important tokens for different inputs~\citep{rao2021dynamicvit,liang2022evit,pan2021ia}, while channel pruning targets searching optimal sub-networks by resorting to both structured and unstructured approaches \citep{chen2021chasing,chavan2022vision,yu2022unified}.

Few-shot knowledge distillation considers knowledge distillation in the context of having only few training samples. Compression methods for deep neural networks typically require finetuning to recover the prediction accuracy when the compression ratio is high. However, conventional finetuning methods are hindered by the need for a large training set. FSKD \citep{li2020few} added a $1\times1$ convolution layer at the end of each layer of the student network and fit the layer-level output features of the student to the teacher. CD~\citep{bai2020few} proposed cross distillation to reduce accumulated errors caused by layer-wise knowledge distillation. In MiR~\cite{wang2022compressing}, the teacher's layers before the penultimate are replaced with well-tuned compact ones. Unlike previous methods that adopt filter-level pruning, PRACTISE~\citep{wang2023practical} was built on block-level pruning and can recover the network's accuracy using only tiny sets of training images. 

Since few-shot compression of ViT has not been explored, while the current SOTA methods for ViT compression always falter in the few-shot context (as will be shown by our experiments), in this paper, we mainly compare our DC-ViT with the SOTA few-shot CNN pruning method.

\section{Method}
\label{sec:method}

Now we propose Dense Compression of Vision Transformers (DC-ViT). The overall framework is illustrated in \cref{fig: DC-ViT framework}. Our strategy can be divided into three primary stages: 1) determine the compressed model's network structure based on the target compression ratio; 2) a block-trial procedure to find recoverability scores for different modules, and generate a synthetic metric set to select blocks with highest recoverability; 3) a progressive approach to prune and finetune the model using few-shot and unlabeled training set. 

\subsection{Determining the Network Structure}
\label{sec: structure determination}

For few-shot learning, removing one entire block leads to a change that can be well approximated by a linear layer (which can be absorbed)~\citep{wang2023practical}. Furthermore, block pruning leads to significantly higher throughput than filter pruning, thus it is beneficial to remove blocks completely. Nevertheless, while the removal of a single block can be \emph{approximately} compensated by linear adjustments, the cumulative errors from dropping multiple blocks can escalate to a non-linear magnitude. Hence, we propose to \emph{replace a dropped block with a small MLP} to remedy such errors.

The structure of ViT makes our remedy particularly suitable. As pointed out by~\cite{wang2023RIFormer}, the attention module takes up roughly 46\% of a Transformer block's latency. After we remove the attention module, the remaining part of a block is exactly an MLP; by pruning this MLP to a smaller one (as in the left side of \cref{fig: DC-ViT framework}), we can not only \emph{achieve dense compression}, but also \emph{reuse parameters} from the original MLP. Our experiments highlighted the significance of parameter reusing in few-shot compression (\cf \cref{table: ablation studies}). Therefore, in our \emph{dense} compression process, we need to determine two things based on the target MACs: the number of blocks to remove and the extent to which the MLPs within these blocks should be reduced.

We calculate the minimum number of blocks that should be compressed to meet the target, denoted as $k$, \ie, 
\begin{equation}
\label{eq: k}
k = \left\lceil \frac{\mathrm{MACs}_o - \mathrm{MACs}_p}{\mathrm{MACs}_a + \mathrm{MACs}_m} \right\rceil \,,
\end{equation}
where $\mathrm{MACs}_o$ and $\mathrm{MACs}_p$ respectively denote MACs of the origin and pruned models ($\mathcal{M}_O$ and $\mathcal{M}_P$), $\mathrm{MACs}_a$ denotes MACs of the attention module \emph{including} the corresponding layer norm, while $\mathrm{MACs}_m$ refers to MACs of the MLP module \emph{excluding} the corresponding layer norm of one ViT block, respectively. $\lceil\cdot\rceil$ is the ceiling function. 

\noindent\textbf{Minimize the number of compressed blocks.} As shown in \cref{fig: MACs-options}, one target MACs can be obtained by removing different numbers of blocks. We can choose to prune more blocks but retain a larger ratio of their corresponding MLPs. Alternatively, we can also prune fewer blocks but keep a smaller part of the MLPs in these blocks. For example, to keep 85\% of the original model's MACs, we can throw away 2, 3, or 4 blocks, as shown in \cref{fig: MACs-options}. However, the color bars, which represent the accuracy of the compressed models, clearly indicate that minimizing the number of compressed blocks is preferable when multiple compression configurations are feasible.

To further simplify our method, we allocate the \emph{same} drop ratio into every MLP module of the $k$ compressed blocks. Hence, the drop ratio of MLP nodes $r_d$ will be
\begin{equation}
r_d = \frac{\mathrm{MACs}_o - \mathrm{MACs}_p - k \cdot \mathrm{MACs}_a}{k \cdot \mathrm{MACs}_m} \,.
\end{equation}
The MLP in a ViT block has 2 fully connected (FC) layers, with the dimension of features changing from $d$ to $4d$ and then back to $d$ again. We change the dimensionality of the intermediate result from $4d$ to $4d(1-r_d)$, which finalizes the compression structure. Our ablation studies show that randomly picking $4d(1-r_d)$ nodes out of $4d$ works surprisingly well, thus DC-ViT adopts this simple random strategy.

\subsection{Block Selection}

In order to select which blocks are to be compressed, we adopt a block-wise trial procedure to obtain candidate models. Next, we generate synthetic images as a tiny metric set to evaluate the recoverability of different blocks after compression.

\subsubsection{Block-wise Trial}

Once the structure parameters ($k$ and $r_d$) are determined, we operate a block-wise trial procedure to obtain candidate models $\mathcal{M}_{P\left(\mathcal{B}_i\right)}$ with block $\mathcal{B}_i$ compressed, then use the criterion proposed in the following section to calculate the recoverability scores of different blocks.

For a few-shot task with $C$ classes, let $\mathcal{D}_{\mathcal{T}}$ denote the tiny subset of the original training set \emph{without} labels. In some applications, not only is the complete training set unavailable, but also labels for the accessible tiny set. Therefore, unsupervised few-shot compression is preferred. 

In the block-wise trial step, we obtain candidate models $\mathcal{M}_{P(\mathcal{B}_i)}$ with block $\mathcal{B}_i$ replaced by a small MLP, as described in \cref{sec: structure determination}. Following MiR~\citep{wang2022compressing}, to finetune $\mathcal{M}_{P(\mathcal{B}_i)}$, we use the mean squared error (MSE) loss for feature mimicking to minimize the feature gap between the pruned and original models, \ie,
\begin{equation}
    \label{eq:l_mse}
    \mathcal{L}_{MSE}\left(\mathcal{M}_{O}, \mathcal{M}_{P}, \mathcal{D}_{\mathcal{T}}\right) = \sum_{x \in \mathcal{D}_{\mathcal{T}}} \Vert \mathcal{M}_{O}\left(x\right) - \mathcal{M}_{P}\left(x\right) \Vert_F^2 \,,
\end{equation}
where $\mathcal{M}_{P}\left(x\right)$ and $\mathcal{M}_{O}\left(x\right)$ denote the \emph{union} of all output tokens of the pruned and the original model, respectively. Even though the CLS token is generally considered sufficient as the representation of an image, conducting feature mimicking using all output tokens works \emph{better} than just focusing on the CLS token, as will be shown in \cref{table: ablation studies}. We also find that by \emph{updating only the first few blocks up to the last compressed one}, instead of all blocks, not only improves the accuracy of finetuned models but also reduces the time cost of gradient back-propagation. As shown in \cref{fig: DC-ViT framework}, only the blocks in the yellow color are trainable. Our ablation studies in \cref{table: ablation studies} back up this strategy.

\subsubsection{Generate a Synthetic Metric Set}
\label{sec: synthetic data}

Since the training data is extremely limited in few-shot compression, various few-shot data augmentation methods exist in the few-shot learning domain. Their primary objective is to produce augmented/synthetic images that align with the true data distribution using limited training samples to increase the sample richness. Some rely on hand-crafted rules from experts with specialized domain knowledge~\citep{chen2019image,devries2017improved,kwitt2016one,gao2018low,chu2019spot,alfassy2019laso}, while a surge of probabilistic-based~\citep{li2020dada} and generative-based~\citep{chen2019multi,zhang2019few,schwartz2018delta} approaches have emerged for automatic synthetic image generation with the maturity of meta-learning.

Operating under the assumption that a deep neural network can sufficiently train and retain crucial information from a dataset, we propose to utilize the original pre-trained model to generate synthetic images as the metric set $\mathcal{S}$ for block selection without using external data. As will be illustrated in \cref{sec: block selection}, the metric set is used to reveal the performance of the finetuned candidate models.

Following DeepDream~\citep{mordvintsev2015inceptionism}, we leverage the pre-trained model's intrinsic knowledge to efficiently generate a compact batch of synthetic images, forming a robust metric dataset. Given a batch of Gaussian noise as the initial synthetic images $\mathcal{S} = \{\hat{x}_1, \hat{x}_2, \cdots, \hat{x}_n\}$ and arbitrary target labels $\mathcal{Y} = \{y_1, y_1, \cdots, y_n\}$, the synthetic images are updated by minimizing
\begin{equation}
    \label{eq: synthetic loss}
    \mathcal{L}_{\mathcal{S}} = \min\limits_{\mathcal{S}}\dfrac{1}{|\mathcal{S}|}\sum_{i=1}^{|\mathcal{S}|}\left(\mathcal{L}(\hat{x}_i, y_i) + \mathcal{R}(\hat{x}_i)\right)\,,
\end{equation}
where $\mathcal{L}(\cdot)$ is the cross entropy between the predicted probability and the target label, and $\mathcal{R}(\cdot)$ is the regularization term. We use the image prior regularizer of $\ell_2$ and total variation (TV) proposed in~\citep{mahendran2015understanding}:
\begin{equation}
    \label{eq: regularization}
    \mathcal{R}(\hat{x})=\alpha_{\ell_2} \mathcal{R}_{\ell_2}(\hat{x}) + \alpha_{\mathrm{tv}} \mathcal{R}_{\mathrm{TV}}(\hat{x})\,,
\end{equation}
where $\mathcal{R}_{\ell_2} = \|\hat{x}\|^2$ is utilized to promote image stability and $\mathcal{R}_{\mathrm{TV}}=\sum_{i, j}\left(\left(\hat{x}_{i, j+1}-\hat{x}_{i j}\right)^2+\left(\hat{x}_{i+1, j}-{\hat{x}}_{i j}\right)^2\right)^{\frac{\beta}{2}}$ can control the sharpness of the synthetic data. 

Since the metric loss $\mathcal{L}_{CE}$ in \cref{eq: metric loss} does not require a class label, the randomly assigned target $y$ in \cref{eq: synthetic loss} only acts as a hint for the pre-trained model to generate synthetic data. After a few iterations, the synthetic images are able to capture the distribution of the original dataset (\cf the appendix).

\subsubsection{Choose the Blocks to Compress}
\label{sec: block selection}

\begin{table}
    \centering
	\small
	\setlength{\tabcolsep}{5pt}
    \begin{tabular}{c|cccc}
    \toprule
    Latency (ms) & Block & Attn & FFN & Attn$+0.5\cdot$FFN \\
    \midrule
    Mean & $102.95$ & $107.70$ & $111.12$ & $103.93$\\
    \midrule
    Std. & $\phantom{00}0.44$ & $\phantom{00}0.11$ & $\phantom{00}0.08$ & $\phantom{00}0.12$\\
    \bottomrule
    \end{tabular}
\caption{Mean and standard deviation of latency in ViT-Base when different parts are removed. ``Attn'': the attention module, ``FFN'': the feed-forward network (MLP), and ``Attn$+0.5\cdot$FFN'': the attention module plus half of MLP.}
\label{table: latency}
\end{table}

Since all ViT blocks in the same model are structurally identical, there is little latency variance caused by compressing different blocks in DC-ViT (\cf \cref{table: latency}). Therefore, the score we propose for block selection is a simple one that evaluates the performance of candidates on the synthetic metric set $\mathcal{S}$ generated as in \cref{sec: synthetic data},
\begin{equation}
    \label{eq: metric loss}
    \begin{aligned}
    &\mathcal{L}_{CE}\left(\mathcal{M}_{O}, \mathcal{M}_{P(\mathcal{B}_i)}, \mathcal{S}\right) =\\
    &- \sum_{x \in \mathcal{S}} \sum_{c=1}^{C} p\left(\mathcal{M}_{O}\left(x\right);c\right) \log \left(p\left(\mathcal{M}_{P(\mathcal{B}_i)}\left(x\right);c\right)\right) \,,
    \end{aligned}
\end{equation}
where $p\left(\mathcal{M}_{P(\mathcal{B}_i)}\left(x\right);c\right)$ denotes the probability of training sample $x$ is classified into the $c$-th class by the $i$-th candidate model, which is calculated using the output CLS token, followed by linear and softmax transformations, akin to $p\left(\mathcal{M}_{O}\left(x\right);c\right)$. 

\begin{figure}
    \centering
    \includegraphics[width=0.475\textwidth]{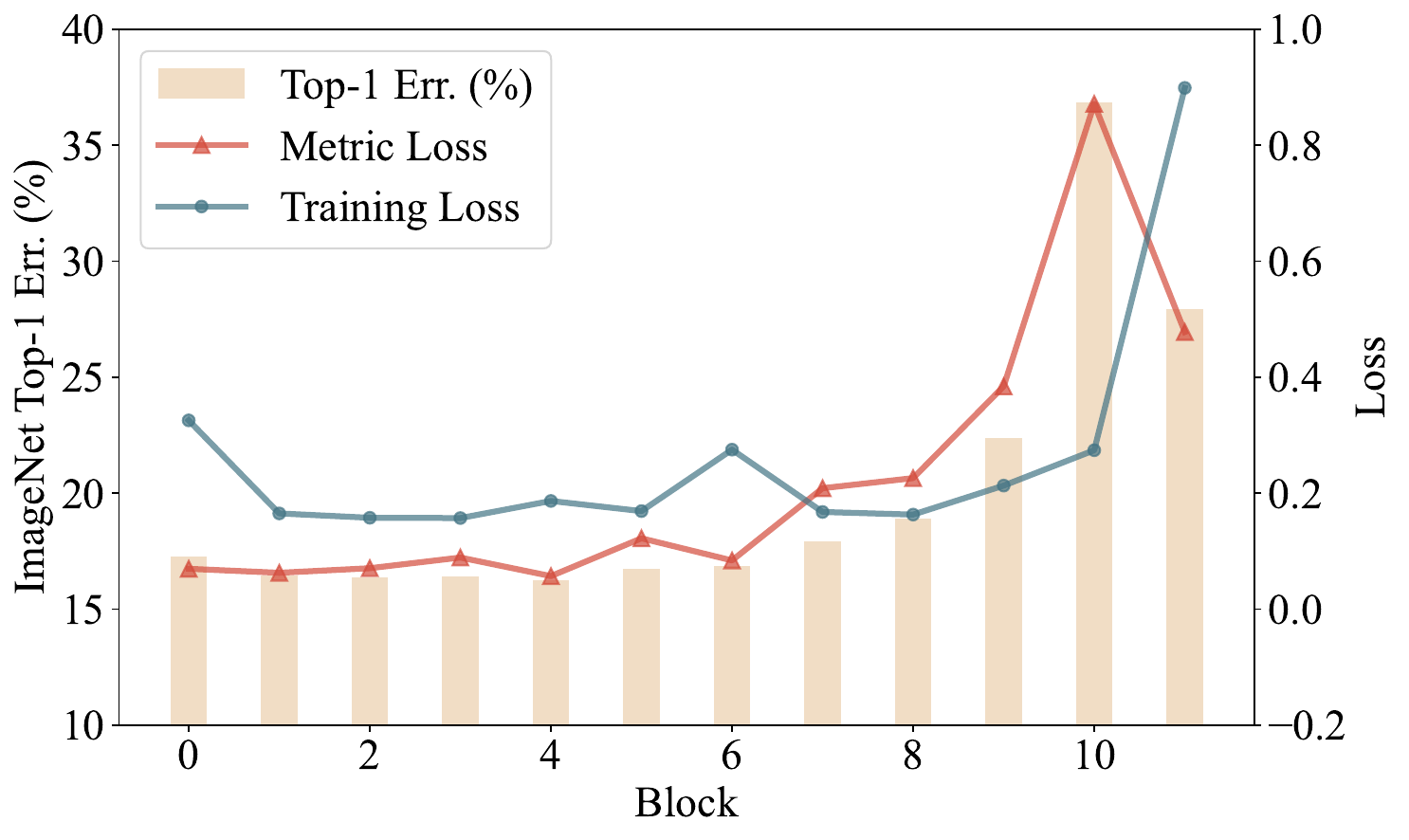}
\caption{The top-1 error of 12 candidate models (by compressing 12 ViT blocks one at a time) on the test dataset and different criteria for block selection. The top-1 error is calculated using the original test set. The metric loss is calculated using the synthetic metric set $\mathcal{S}$. The training loss is calculated using the tiny training set $\mathcal{D}_{\mathcal{T}}$. We scaled them to the same range.}
\label{fig: alignment}
\end{figure}

\noindent\textbf{Synthetic images bridge the gap between perceived and actual block recoverability.} As illustrated in \cref{fig: alignment}, the top-1 error of candidate models on the test dataset and the metric loss on the synthetic metric set $\mathcal{S}$ are highly correlated. Conversely, the training loss on the tiny training set $\mathcal{D}_{\mathcal{T}}$ shows no such correlation, even though with different data augmentation from the finetuning phase, and this disparity becomes more evident with fewer training samples. We attribute this discrepancy to overfitting on the limited training set. Hence, the use of synthetic images as a metric dataset can address the potential misalignment between the perceived difficulty in model performance recovery (as indicated by the training loss) and the actual performance of finetuned models after removing different blocks.

Evaluating the recoverability directly using the finetuned candidates has two advantages. The performance of candidates consistent with the final structure reflects the recoverability of removing different blocks more intuitively. Furthermore, these candidate models can be reused in subsequent compression without being repeatedly finetuned.

\subsection{Progressive Pruning and Finetuning}

\begin{algorithm}
    \caption{DC-ViT}
    \label{algo: DC-ViT}    
    \begin{algorithmic} [1]
        \floatname{algorithm}{DC-ViT}
        \renewcommand{\algorithmicrequire}{\textbf{Input:}}
        \renewcommand{\algorithmicensure}{\textbf{Output:}}
        \renewcommand{\algorithmiccomment}[1]{\textcolor{ForestGreen}{\hfill$\triangleright$ #1}}
        
        \REQUIRE original model $\mathcal{M}_O$, tiny training set $\mathcal{D}_{\mathcal{T}}$, synthetic metric set $\mathcal{S}$, number of blocks to be compressed $k$, MLP drop ratio  $r_d$
        \ENSURE pruned model $\mathcal{M}_P$
        \STATE $\mathcal{M}_P \gets \mathcal{M}_O$
        \STATE $mList \gets$ empty list
        \FOR {each block $\mathcal{B}_i$ of $\mathcal{M}_P$}
            \STATE $\mathcal{M}_{P\left(\mathcal{B}_i\right)} \gets$ drop the attention and (randomly) $r_d$ of MLP nodes from $\mathcal{B}_i$ \ \ \COMMENT{trial-compress single block}
            \STATE Minimize $\mathcal{L}_{MSE}\left(\mathcal{M}_{O}, \mathcal{M}_{P\left(\mathcal{B}_i\right)}, \mathcal{D}_{\mathcal{T}}\right)$ to update $\mathcal{M}_{P\left(\mathcal{B}_i\right)}$\ \ \COMMENT{finetune corresponding model}
            \STATE $\mathcal{L}_{m} \gets \mathcal{L}_{CE}\left(\mathcal{M}_{O}, \mathcal{M}_{P\left(\mathcal{B}_i\right)}, \mathcal{S}\right)$\ \ \COMMENT{metric loss}
            \STATE Add pair $(\mathcal{L}_{m}, i)$ to $mList$
        \ENDFOR
        \STATE Sort $mList$ based on the metric loss of each block
        \STATE $C \gets$ first $k$ indices of $mList$\ \ \COMMENT{blocks to compress}
        \STATE Progressively compress $\mathcal{B}_{C[:]}$ in $\mathcal{M}_P$\ \ \COMMENT{line 5-6}
        \RETURN $\mathcal{M}_P$
    \end{algorithmic}
\end{algorithm}

Our findings indicate that to compress more than one block, progressive compression---compressing blocks sequentially, one after the other---is more effective. Since few-shot compression is a delicate process, it's beneficial to minimize extensive modifications to the model structure in a single stage. Adopting a gradual approach to compression not only mitigates potential risks but also demonstrates superior performance, as will be shown in \cref{table: ablation studies}.

We provide the pseudo-code for DC-ViT in \cref{algo: DC-ViT}. In the beginning, we determine the compressed network's structure. Then, we sequentially trial-removed different blocks in the original model $\mathcal{M}_O$ to create candidates $\mathcal{M}_{P\left(\mathcal{B}_i\right)}$ and finetune the candidates with the tiny training set $\mathcal{D}_{\mathcal{T}}$. We generate a synthetic metric set to evaluate the rank of blocks' recoverability. Finally, we select the $k$ blocks with the smallest metric values and progressively compress the selected blocks in $\mathcal{M}_O$ to obtain the pruned model $\mathcal{M}_P$.

\section{Experiments}
\label{sec:experiments}

In this section, we report results across various architectures, followed by ablation studies to validate the efficacy of our \emph{dense} compression method. Additionally, we offer some suggestions for few-shot ViT compression.

\subsection{Experimental Settings}

\textbf{Architectures.} We used the vanilla ViT models initially introduced by~\citet{dosovitskiy2021an}, the smaller and the larger ViT variants proposed by~\citet{touvron2021training}, DeiT-B~\citep{touvron2021training} and Swin-B~\citep{liu2021Swin}. We also tested our method on CNN architectures, using ResNet-34 and MobileNetV2 as the backbone. Finally, we demonstrate the transferability of the compression models on downstream tasks.

\noindent\textbf{Benchmark data.} We randomly sampled tiny training sets from ImageNet-1k~\citep{ILSVRC15} and evaluated the top-1 / 5 accuracy of the pruned models on the validation set. When we reported the mean accuracy with the standard deviation, we ran the experiments 5 times with 5 random seeds (2021 $\sim$ 2025); otherwise, we took 2023 as the random seed.

\begin{table}
    \centering
	\small
    \begin{tabular}{lclclc}
    \toprule
    Batch size & $64$ & Epochs & $2000$ \\
    Warmup epochs & $20$ & Weight decay & $1e-4$ \\
    Optimizer & AdamW & Learning rate & $3e-5$ \\
    Learning rate decay & cosine & Gradient Clip. & $1.0$ \\
    Drop path rate & $0.5$ & Drop rate & $0.0$ \\
    Rand Augment & $9 / 0.5$ & Color Jitter & $0.3$\\
    \bottomrule
    \end{tabular}
\caption{Hyper-parameters of our DC-ViT.}
\label{table: hyper-parameters}
\end{table}

\noindent\textbf{Training procedure.} \cref{table: hyper-parameters} displays the default hyper-parameter setting we used at training time for all our experiments. Unless otherwise specified, the experiments were conducted at $224^2$ resolution.

\subsection{Baseline Pruning Methods}
\label{sec: baseline}
To the best of our knowledge, the field of few-shot compression for ViTs is currently unexplored. Previous ViT compression methods always rely on the complete ImageNet-1k dataset and often falter in the few-shot context. For example, S$^2$PViT~\citep{chen2021chasing}, EViT~\citep{liang2022evit}, are the SOTA ViT compression methods that achieve excellent performance on DeiT. However, when we applied them to a few-shot setting and used 1000 training samples (which is already a large training set in few-shot compression), \emph{neither of them converged}.

Moreover, most methods for few-shot compression of CNNs involve specific operations on convolutional layers, which are not easily transferable to ViT architectures. Therefore, we adopt the SOTA few-shot CNN pruning method, PRACTISE~\citep{wang2023practical}, which can be directly extended to ViT as a strong baseline. To further verify the effectiveness of our DC-ViT, we also extend DC-ViT to CNN architectures and benchmark its performance against notable few-shot CNN pruning methods, including FSKD~\citep{yin2020dreaming}, CD~\citep{bai2020few} and MiR~\citep{wang2022compressing}. When comparing with other few-shot pruning methods, we are mainly concerned about the latency-accuracy tradeoff of the pruned model. The latency of the pruned models is measured using a batch size of $64$ and recorded the average inference time over $500$ runs.

\begin{table*}
  \centering
  \small
  \begin{tabular}{c|l|c|cccc}
    \toprule
    MACs & Methods & Latency (ms) & 50 sample & 100 sample & 500 sample & 1000 sample\\
    \midrule
    \multirow{2}*{\thead{$15.5\times 10^{9}$ \\ ($k=1$, $8.3\%\downarrow$)}} 
    & PRACTISE$^{\ast}$ & $103.2$ ($6.5\%\downarrow$) & $71.28_{\pm 0.24}$ & $75.69_{\pm 6.77}$ & $73.32_{\pm 0.14}$ & $74.71_{\pm 0.13}$ \\
    & DC-ViT-B & ${\bf 101.9}$ (${\bf 8.3\%\downarrow}$) & ${\bf 82.88}_{\pm 0.10}$ & ${\bf 83.30}_{\pm 0.17}$ & ${\bf 83.73}_{\pm 0.14}$ & ${\bf 83.78}_{\pm 0.04}$\\
    \midrule
    \multirow{3}*{\thead{$14.1\times 10^{9}$ \\ ($k=2$, $16.6\%\downarrow$)}}
    & PRACTISE$^{\ast}$  & $94.7$ ($15.4\%\downarrow$) & $69.70_{\pm 0.32}$ & $73.75_{\pm 5.91}$ & $72.54_{\pm 0.31}$ & $74.52_{\pm 0.22}$\\
    & PRACTISE$^{\dag}$ & $94.7$ ($15.4\%\downarrow$) & ${69.87_{\pm 0.48}}$ & $74.11_{\pm 6.01}$  & $72.74_{\pm 0.12}$ & $74.69_{\pm 0.10}$\\
    & DC-ViT-B & ${\bf 93.5}$ (${\bf 16.7\%\downarrow}$) &  ${\bf 80.59}_{\pm 0.17}$  & ${\bf 81.58}_{\pm 0.09}$ & ${\bf 82.74}_{\pm 0.06}$ & ${\bf 83.15}_{\pm 0.08}$\\
    \midrule
    \multirow{3}*{\thead{$12.7\times 10^{9}$ \\ ($k=3$, $24.8\%\downarrow$)}} 
    & PRACTISE$^{\ast}$ & $85.9$ ($23.5\%\downarrow$) &  $63.75_{\pm 0.44}$  & $67.58_{\pm 4.42}$ & $70.03_{\pm 0.56}$ & $72.91_{\pm 0.04}$\\
    & PRACTISE$^{\dag}$ & $85.9$ ($23.5\%\downarrow$) & $65.07_{\pm 0.49}$  & $69.60_{\pm 4.96}$  & $70.54_{\pm 0.12}$ & $73.28_{\pm 0.03}$ \\
    & DC-ViT-B & ${\bf 84.9}$ (${\bf 24.5\%\downarrow}$) & ${\bf 74.70}_{\pm 0.81}$ & ${\bf 77.03}_{\pm 0.26}$ & ${\bf 80.18}_{\pm 0.37}$ & ${\bf 81.60}_{\pm 0.10}$\\
    \bottomrule
  \end{tabular}
\caption{Top-1 validation accuracy (\%) on ImageNet-1k for pruning ViT-Base with 50, 100, 500, 1000 samples. $^{\ast}$ means the method is implemented by us, $^{\dag}$ means the epoch was set to $2000 \cdot k$ ($k$ is the number of compressed blocks) for a fair comparison. $\downarrow$ means the percentage of reduction. The accuracy and latency of the original ViT-Base are $84.41\%$ and 111.1 ms, respectively.}
\label{table: ViT-Base}
\end{table*}

\subsection{DC-ViT Performance}
We demonstrate our DC-ViT performance on various ViT architectures, including plain ViT \citep{dosovitskiy2021an}, DeiT \citep{touvron2021training} and Swin \citep{liu2021Swin}, as well as on CNN architectures. 
\subsubsection{Results on ViT-Base}
\label{sec: ViT-Base}
We primarily compared our DC-ViT method with the SOTA few-shot CNN pruning method PRACTISE~\citep{wang2023practical} on ViT-B. 

\noindent\textbf{Implementation details.} We set the MACs of the pruned models in DC-ViT to be equal to the cases where the baseline drops $1$, $2$, or $3$ entire blocks respectively, and the number of training samples was set to $50$, $100$, $500$ and $1000$. Moreover, we ran the baseline for $2000 \cdot k$ and $2000$ epochs respectively for a fair comparison, because DC-ViT takes a progressive compression strategy and the finetuning step costs $2000$ epochs in each iteration.

As shown in \cref{table: ViT-Base}, under the same MACs settings, compared with removing complete blocks, our \emph{dense} compression method by dropping the attention module and part of the MLP module always gains lower latency, especially when the number of compressed blocks and training samples increases, and our DC-ViT approach outperforms the baseline by a significant margin, achieving roughly 10 percentage points higher top-1 accuracy.

\subsubsection{Results on CNN Architectures}
\begin{table}
    \centering
    \small
    \begin{tabular}{l|cc}
    \toprule
    Methods & Latency (ms) & Top-1 / 5 acc (\%) \\
    \midrule
    ResNet-34 & \textcolor{c2}{$34.4$} & \textcolor{c2}{$73.3 / 91.4$} \\
    \midrule[0.3pt]
    FSKD~\citep{li2020few} & $29.0 (15.8\%\downarrow)$ & $45.3 / 71.5$ \\
    CD~\citep{bai2020few} & $29.0 (15.8\%\downarrow)$ & $56.2 / 80.8$ \\
    MiR~\citep{wang2022compressing} & $29.0 (15.8\%\downarrow)$ & $64.1 / 86.3$ \\
    PRACTISE~\citep{wang2023practical} & $29.0 (15.8\%\downarrow)$ & $70.3 / 89.6$ \\
    {\bf Ours} & ${\bf 28.8 (16.5\%\downarrow)}$ & ${\bf 72.2 / 90.7}$\\
    \midrule
    MobileNetV2 & \textcolor{c2}{$23.2$} & \textcolor{c2}{$71.9 / 90.3$} \\
    \midrule[0.3pt]
    FSKD~\citep{li2020few} & $19.4 (16.2\%\downarrow)$ & $48.4 / 73.9$ \\
    MiR~\citep{wang2022compressing} & $19.4 (16.2\%\downarrow)$ & $67.6 / 87.9$ \\
    PRACTISE~\citep{wang2023practical} & ${\bf 18.8 (19.1\%\downarrow)}$ & $69.3 / 88.9$ \\
    {\bf Ours} & ${\bf 18.8 (19.1\%\downarrow)}$ & ${\bf 69.8 / 89.2 }$\\
    \bottomrule
    \end{tabular}
\caption{Top-1 validation accuracy (\%) on ImageNet-1k for pruning Resnet-34 and MobileNetV2. Previous few-shot pruning methods and our DC-ViT are used to prune Resnet34 and MobileNetV2 with 50 training samples.}
\label{table: cnn-methods}
\end{table}

Few-shot compression on ViT remains under-explored in existing literature, thus we sought to further validate the efficacy of our approach by extending it to CNN architectures and benchmark its performance against few-shot CNN pruning methods FSKD~\citep{yin2020dreaming}, CD~\citep{bai2020few}, MiR~\citep{wang2022compressing}. We used ResNet-34 and MobileNetV2 as the backbone and use $50$ training samples. Since the structure of CNN doesn't facilitate the reuse of MLP nodes for \textit{dense} compression like DC-ViT, we tested the cases where our method eliminates entire blocks. As shown in \cref{table: cnn-methods}, DC-ViT still outperforms previous methods by a significant margin.

\subsubsection{Results on Various ViT Variants}

\begin{table}
    \centering
    \small
	\setlength{\tabcolsep}{2pt}
    \begin{tabular}{l|c|ccc}
    \toprule
    Model & Methods & Latency (ms)  & Top-1 / 5 acc (\%) \\
    \midrule
    \multirow{3}*{ViT-T} 
    & Original & \textcolor{c2}{$20.5$} & \textcolor{c2}{$75.33$ / $92.81$} \\
    & PRACTISE$^{\dag}$ & $17.4$ ($15.1\%\downarrow$) & $54.88$ / $82.43$\\
    & DC-ViT-T & ${\bf 17.2}$ (${\bf 16.5\%\downarrow}$) & ${\bf 64.42}$ / ${\bf 86.65}$\\
    \midrule[0.3pt]
    \multirow{3}*{ViT-S}
    & Original & \textcolor{c2}{$42.9$} & \textcolor{c2}{$81.38$ / $96.07$}\\
    & PRACTISE$^{\dag}$ & $36.5$ ($14.9\%\downarrow$) & $69.18$ / $90.42$\\
    & DC-ViT-S & ${\bf 36.1}$ (${\bf 16.7\%\downarrow}$) & ${\bf 78.64}$ / ${\bf 94.79}$ \\
    \midrule[0.3pt]
    \multirow{3}*{ViT-L}
    & Original & \textcolor{c2}{$341.9$} & \textcolor{c2}{$85.68$ / $97.74$}\\
    & PRACTISE$^{\dag}$ & $317.0$ ($7.3\%\downarrow$) & $83.99$ / $97.12$\\
    & DC-ViT-L & ${\bf 314.3}$ (${\bf 8.2\%\downarrow}$) & ${\bf 85.49}$ / ${\bf 97.63}$\\
    \midrule[0.3pt]
    \multirow{3}*{DeiT-B} 
    & Original & \textcolor{c2}{$110.2$} & \textcolor{c2}{$81.74$ / $95.58$}\\
    & PRACTISE$^{\dag}$ & $94.2$ ($14.5\%\downarrow$) & ${79.30}$ / ${94.31}$ \\
    & DC-DeiT-B & ${\bf 93.6}$ (${\bf 16.7\%\downarrow}$) & ${\bf 81.26}$ / ${\bf 95.35}$ \\
    \midrule[0.3pt]
    \multirow{3}*{Swin-B}
    & Original & \textcolor{c2}{$155.9$} & \textcolor{c2}{$84.47$ / $97.46$}\\
    & PRACTISE$^{\dag}$ & $135.6$ ($13.0\%\downarrow$) &  ${82.88}$ / ${96.76}$\\
    & DC-Swin-B & $135.6$ ($13.0\%\downarrow$) & ${\bf 83.82}$ / ${\bf 97.09}$ \\
    \bottomrule
    \end{tabular}
\caption{Top-1 / 5 validation accuracy (\%) on ImageNet-1k for pruning various ViT architectures with 500 samples. $^{\dag}$ means the method is implemented by us. $\downarrow$ means percentage of reduction.}
\label{table: various-arch}
\end{table}

We have further adapted our DC-ViT framework to encompass variants of ViT models, including DeiT-Base and Swin-Base. To ensure a fair and meaningful comparison with PRACTISE~\citep{wang2023practical}, we uniformly set the MACs of the pruned models in DC-ViT to match those in scenarios where PRACTISE drops $2$ entire blocks. However, due to the complex structure of Swin, we directly select and eliminate two pairs of successive Swin blocks, \ie, a total of 4 complete blocks. \cref{table: various-arch} demonstrates that DC-ViT consistently outperforms other methods across all these ViT architectures.

\subsection{Ablation for Few-shot ViT Compression}

\begin{table}
    \centering
	\small
	\setlength{\tabcolsep}{4pt}
    \begin{tabular}{l|cc}
    \toprule
    MACs & $15\%\downarrow$ ($k=2$) &  $30\%\downarrow$ ($k=4$)  \\
    \midrule
    DC-DeiT-B & ${\bf 83.25}$ & ${\bf 79.28}$ \\
    \midrule[0.3pt]
    top-$k$ (2000 epochs) & $82.84$ & $67.95$ \\ 
    top-$k$ ($2000\cdot k$ epochs) & $83.05$ & $78.50$\\
    \midrule[0.3pt]
    finetune the entire ViT & $82.66$ & $78.17$ \\
    \midrule[0.3pt]
    compress ($k+1$) blocks & $82.84$ & $75.07$ \\
    \midrule[0.3pt]
    re-initialize MLP weights & $82.60$ & $77.29$ \\
    GeLU-based weights reuse & $83.07$ & $79.25$ \\
    \midrule[0.3pt]
    metric with MSE loss & $74.02$ & $69.22$ \\
    \midrule[0.3pt]
    mimicking on class token & $82.73$ & $78.33$ \\
    \bottomrule
    \end{tabular}
\caption{Ablation of alternative settings in DC-ViT. We pruned ViT-Base with 500 samples to achieve $15\%$ and $30\%$ MACs reduction respectively.}
\label{table: ablation studies}
\end{table}

We designed ablation studies to provide further insights into our framework. The results are shown in \cref{table: ablation studies}, based on which we provide some suggestions for compressing ViT when sample availability is limited.

\noindent\textbf{Use progressive pruning procedure.} As the top-$k$ compression strategy, we finetuned for $2000$ and $2000\cdot k$ epochs respectively for a fair comparison. Either way, the progressive compression strategy achieves higher accuracy, especially when more blocks are compressed.

\noindent\textbf{Update the front part of blocks.} When finetuning the entire ViT, the accuracy decreased in all compression settings, especially when more blocks are compressed, which means our partial finetuning approach is not only efficient but also effective in few-shot contexts.

\noindent\textbf{Reuse MLP weights.} If the small MLP's weights are not reused but randomly re-initialized for finetuning, we can find that the more blocks are compressed, the more accuracy drops. In addition, we also tried some other ways of selecting the MLP nodes to be reused, such as reusing weights based on the mean activation value after GeLU. However, the GeLU-based method did not show any obvious advantage over random selection.

\noindent\textbf{Compress as few blocks as possible.} Given a specific compression ratio, if various options are available, the accuracy tends to decrease more drastically when more blocks are compressed. Therefore, we believe that it is better to compress as few blocks as possible so long as the target compression ratio is satisfied, which is also evidenced by \cref{fig: MACs-options}.

We also tried some other alternative settings, such as using MSE loss to choose candidate models and conducting feature mimicking on the CLS token, but none of them achieved better performance than our default settings.

\subsection{Transfer Learning to Downstream Tasks}

\begin{table}
    \centering
	\small
	\setlength{\tabcolsep}{4pt}
    \begin{tabular}{l|cccccc}
    \toprule
    Model & \rotatebox{90}{CIFAR-10} & \rotatebox{90}{CIFAR-100} & \rotatebox{90}{Flowers} & \rotatebox{90}{Pets} & \rotatebox{90}{CUB-200} & \rotatebox{90}{Indoor67}\\
    \midrule
    EfficientNet-B7~\citep{tan2019efficientnet} & 98.9 & 91.7 & 98.8 & - & - & - \\
    ViT-B~\citep{dosovitskiy2021an} & $97.4$ & $88.0$ & $98.5$ & $92.9$ & $84.1$ & $83.1$\\
    DC-ViT-B & $98.0$ & $87.5$ & $99.3$ & $92.2$ & $83.5$ & $82.7$\\
    \bottomrule
    \end{tabular}
\caption{Transfer learning tasks' performance of DC-ViT-B with $24.8\%$ latency reduction after few-shot compression. We report the top-1 accuracy of finetuning the original ViT-B model and our DC-ViT-B on downstream classification datasets.}
\label{table: downstream}
\end{table}
Finally, we evaluate the generalizability of our DC-ViT.

\noindent\textbf{Implementation details.} We finetuned our DC-ViT-B model compressed in \cref{sec: ViT-Base} on a collection of commonly used recognition datasets: CIFAR-10 and CIFAR-100~\citep{krizhevsky2009learning}, Flowers~\citep{nilsback2006visual}, Oxford-IIIT Pets~\citep{parkhi2012cats}, CUB-200~\citep{wah2011caltech} and Indoor67~\citep{quattoni2009recognizing}. During the finetuning phase, an AdamW optimizer was adopted with batch size 256, learning rate $5\times 10^{-4} \times\frac{\text{batchsize}}{512}$.

\cref{table: downstream} shows the results of the Top-1 accuracy of the original ViT-B and the pruned DC-ViT-B on downstream datasets. Compared to the SOTA ConvNets and transformer-based models, the pruned model achieved comparable or even better performance on all downstream classification tasks, which shows that the efficiency
demonstrated on ImageNet can be preserved on downstream tasks by our few-shot compression method.

\section{Conclusions}

In this paper, we proposed a novel framework, Dense Compression of Vision Transformers (DC-ViT), for few-sample compression on Vision Transformers, which achieves not only much denser compression options in terms of MACs by dropping the attention module and adjusting the MLP drop ratio but also a better performance of the pruned models with higher accuracy and lower latency.

\textbf{Limitations and Future Works.} 
One common limitation of DC-ViT and other few-shot compression methods is that the ratio of compressed MACs cannot be too large (\eg, mostly $\le 30\%$), otherwise, the compressed model's accuracy will drop dramatically. Since DC-ViT introduces small MLPs into pruned blocks, it has the potential to partially solve this difficulty in the future. 
Another limitation is that few-shot compression is mainly restricted to recognition. However, tasks like detection and segmentation are more intriguing. We plan to test the effect of DC-ViT on ViT-based detector compression in future works.

\section*{Acknowledgements}
We thank Professor Jianxin Wu for the guidance and help with the experiments and manuscript. 
This research was partly supported by the National Natural Science Foundation of China under Grant 62276123 and Grant 61921006.

{
    \small
    \bibliographystyle{ieeenat_fullname}
    \bibliography{main}
}

\clearpage
\appendix
\section*{Appendix}

In this appendix, we present additional details and results to complement the main paper. We first provide the full results of the latency analysis conducted on the 12 blocks of the ViT-Base model. Subsequently, we illustrate visualizations of synthetic metric collections generated by various iterations of the ViT model, alongside the duration required for updating synthetic images.

\section{Latency analysis}
We conducted latency tests on the 12 blocks of the ViT-Base model, analyzing the impact of completely or partially removing each block. The term ``Attn$+0.5\cdot$FFN'' denotes the standard approach of DC-ViT in compressing a ViT block, which involves eliminating the entire attention module and a portion of the MLP module. To ensure the robustness of our results, we performed the latency tests five times utilizing five distinct random seeds (ranging from 2021 to 2025) and reported the average latency along with the standard deviation. The full results are detailed in \cref{table: latency_sup}. Owing to the consistent architectural design of all ViT blocks within the same model, our findings indicate minimal latency variance when different blocks are compressed using the DC-ViT approach.

\begin{table}
    \centering
    \setlength{\tabcolsep}{10pt}
    \begin{tabular}{c|cc}
    \toprule
    Latency (ms) & Block & Attn \\
    \midrule
    0 & $102.29_{\pm 0.19}$ & $107.50_{\pm 0.26}$ \\
    1 & $102.85_{\pm 0.03}$ & $107.64_{\pm 0.14}$ \\
    2 & $102.81_{\pm 0.19}$ & $107.69_{\pm 0.08}$ \\
    3 & $102.89_{\pm 0.14}$ & $107.71_{\pm 0.08}$ \\
    4 & $102.88_{\pm 0.23}$ & $107.53_{\pm 0.14}$ \\
    5 & $103.00_{\pm 0.06}$ & $107.66_{\pm 0.10}$ \\
    6 & $102.97_{\pm 0.12}$ & $107.74_{\pm 0.07}$ \\
    7 & $102.89_{\pm 0.12}$ & $107.73_{\pm 0.06}$ \\
    8 & $102.82_{\pm 0.16}$ & $107.87_{\pm 0.07}$ \\
    9 & $102.84_{\pm 0.05}$ & $107.80_{\pm 0.19}$ \\
    10 & $102.96_{\pm 0.18}$ & $107.82_{\pm 0.15}$ \\
    11 & $102.84_{\pm 0.23}$ & $107.68_{\pm 0.10}$ \\
    \bottomrule
    \toprule
    Latency (ms) & FFN & Attn$+0.5\cdot$FFN \\
    \midrule
    0 & $110.99_{\pm 0.36}$ & $103.84_{\pm 0.22}$\\
    1 & $111.18_{\pm 0.06}$ & $103.97_{\pm 0.18}$\\
    2 & $111.13_{\pm 0.12}$ & $104.04_{\pm 0.09}$\\
    3 & $111.14_{\pm 0.23}$ & $104.03_{\pm 0.08}$\\
    4 & $111.15_{\pm 0.07}$ & $104.00_{\pm 0.13}$\\
    5 & $111.23_{\pm 0.06}$ & $104.09_{\pm 0.09}$\\
    6 & $111.06_{\pm 0.23}$ & $103.98_{\pm 0.06}$\\
    7 & $111.10_{\pm 0.11}$ & $103.92_{\pm 0.10}$\\
    8 & $111.02_{\pm 0.20}$ & $103.74_{\pm 0.40}$\\
    9 & $111.07_{\pm 0.16}$ & $103.72_{\pm 0.51}$\\
    10 & $111.20_{\pm 0.06}$ & $103.88_{\pm 0.19}$\\
    11 & $111.19_{\pm 0.02}$ & $103.89_{\pm 0.28}$\\
    \bottomrule
    \end{tabular}
    \caption{The latency in milliseconds (ms) of ViT-Base when different parts are removed across the 12 blocks. ``Attn'': the attention module, ``FFN'': the feed-forward network (MLP), and ``Attn$+0.5\cdot$FFN'': the attention module plus half of MLP.}
    \label{table: latency_sup}
\end{table}

\section{Synthetic Metric Sets}
\label{sec: synthetic images}
\begin{figure}
    \centering
    \includegraphics[width=\linewidth]
    {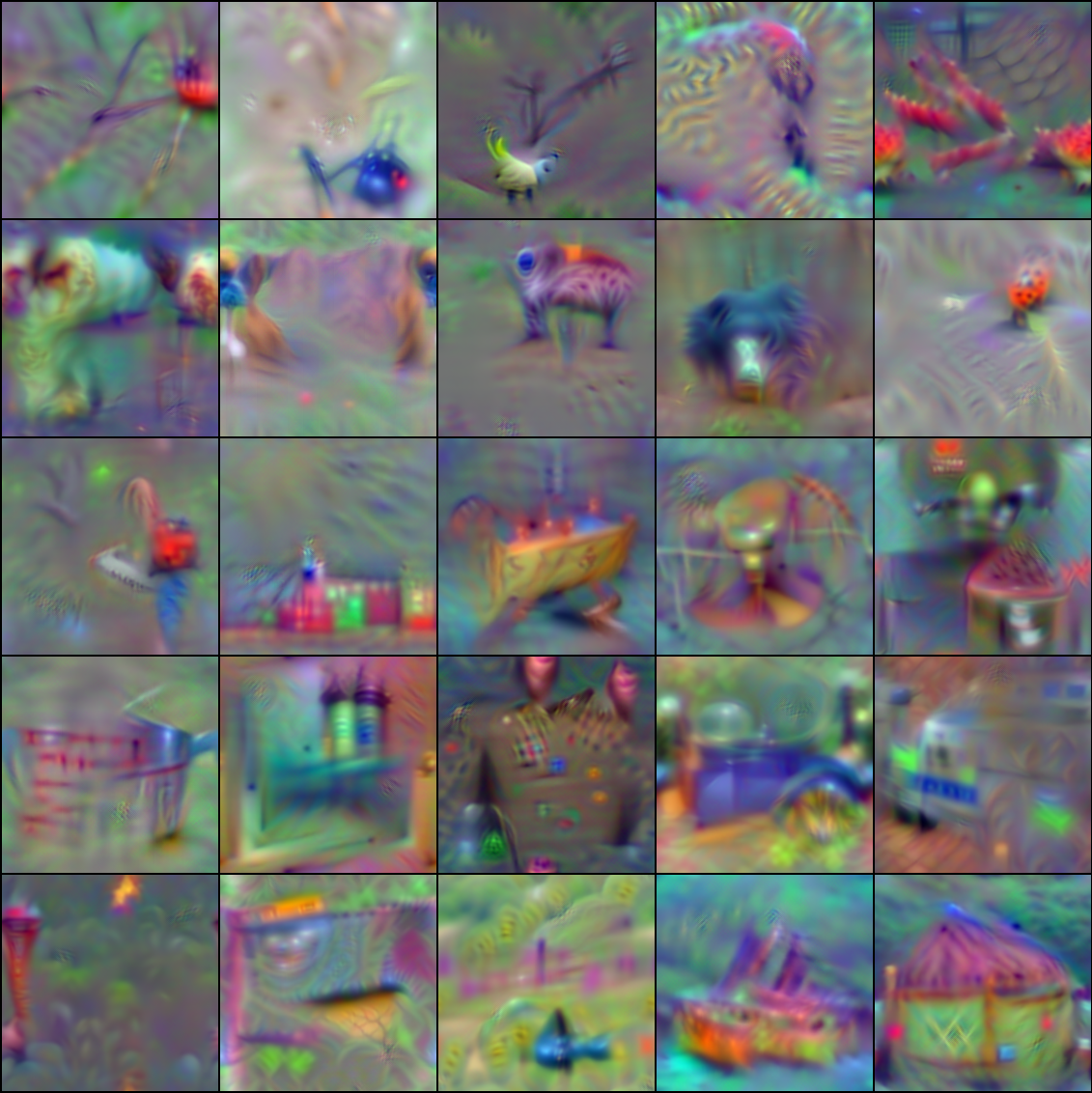}
    \caption{The synthetic metric set generated by ViT-Tiny.}
    \label{fig: vit-tiny}
\end{figure}

\begin{figure}
    \centering
    \includegraphics[width=\linewidth]
    {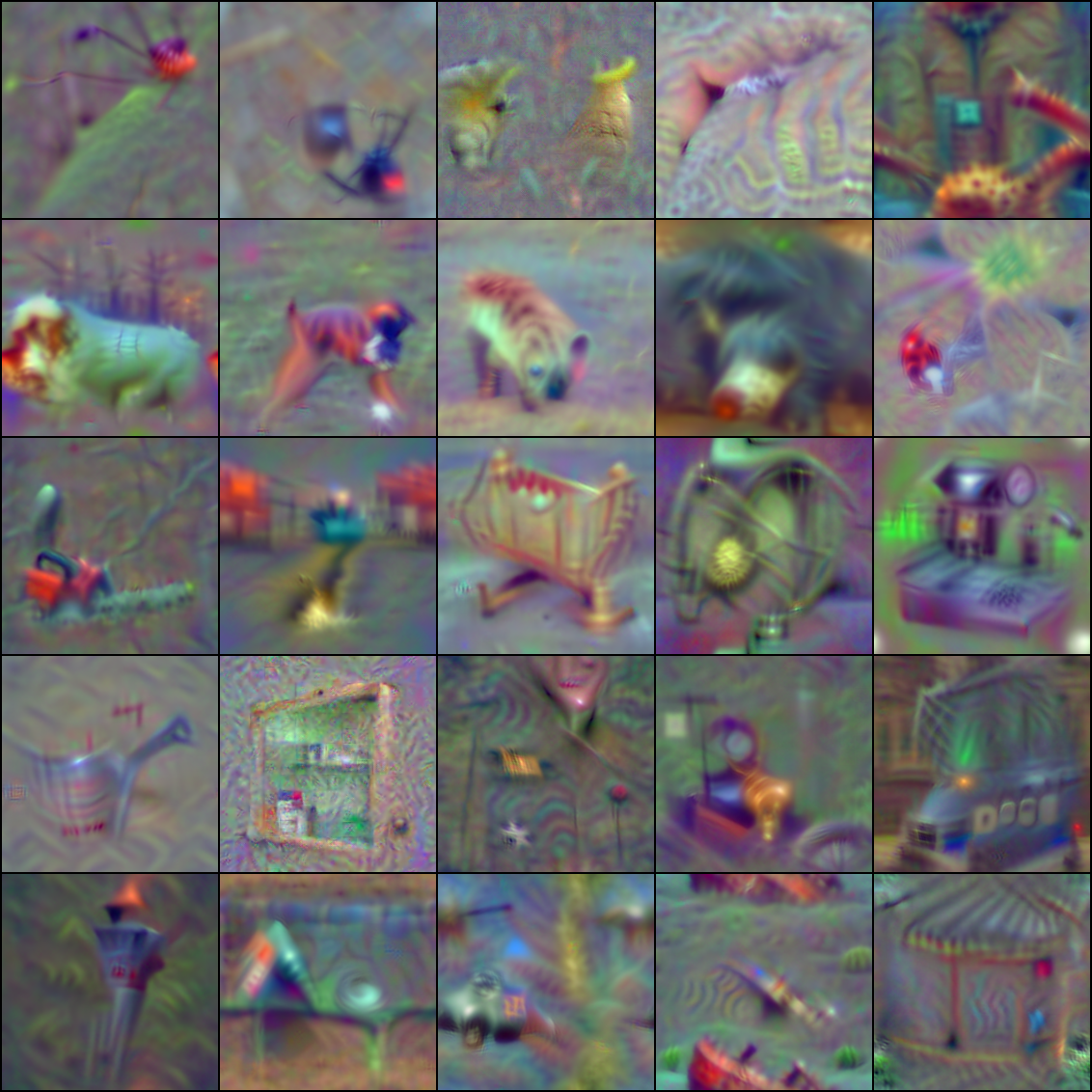}
    \caption{The synthetic metric set generated by ViT-Small.}
    \label{fig: vit-small}
\end{figure}

\begin{figure}
    \centering
    \includegraphics[width=\linewidth]
    {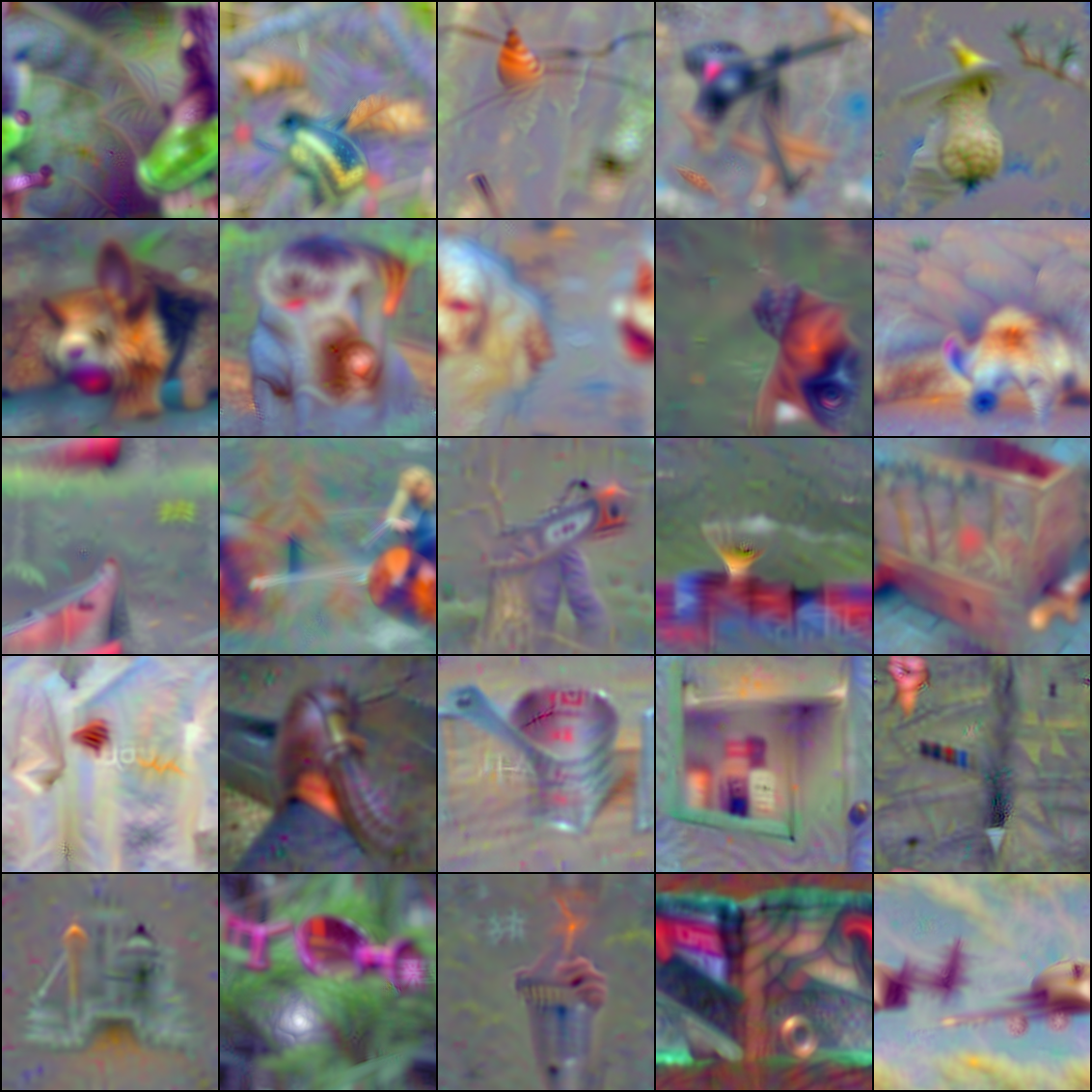}
    \caption{The synthetic metric set generated by ViT-Base.}
    \label{fig: vit-base}
\end{figure}

\begin{figure}
    \centering
    \includegraphics[width=\linewidth]
    {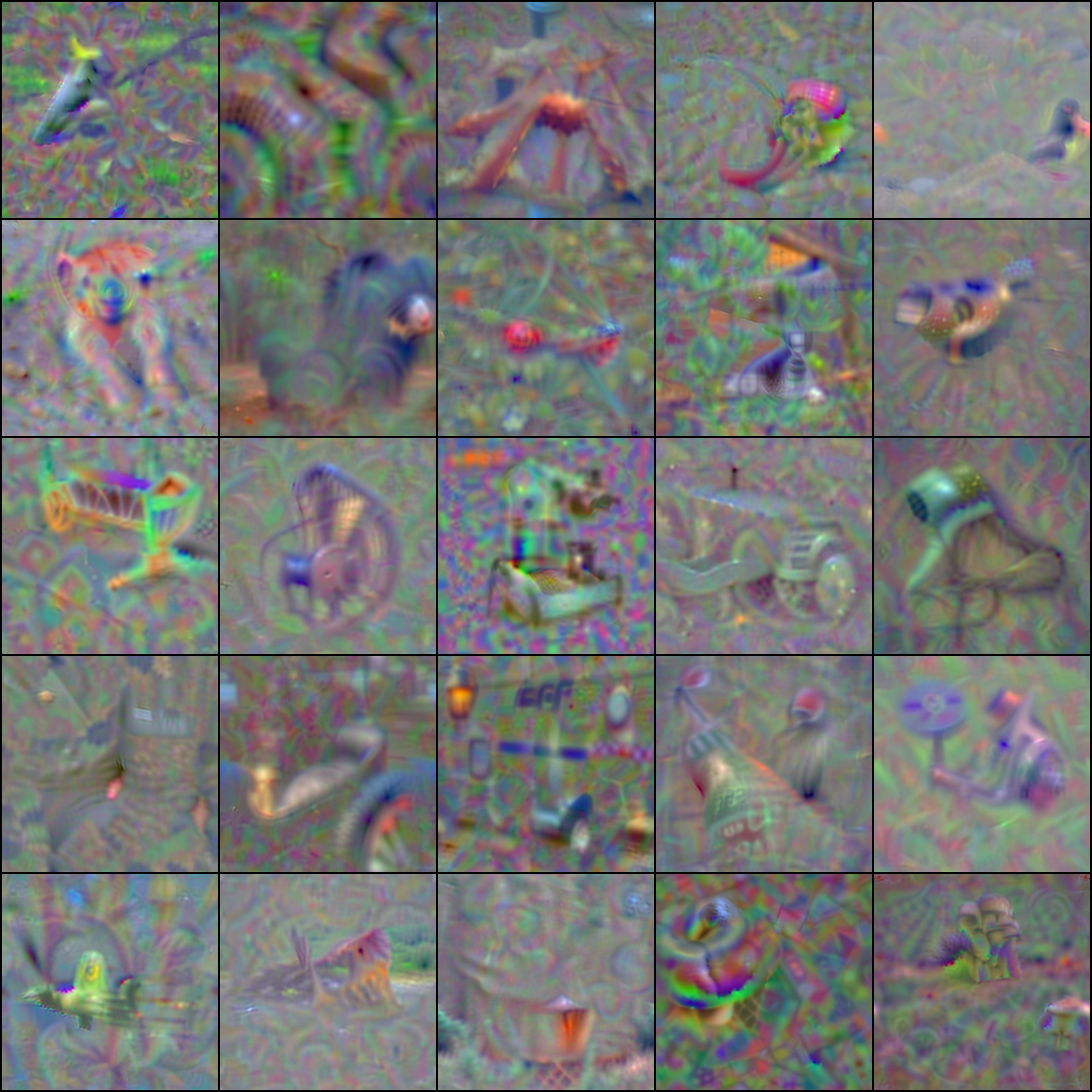}
    \caption{The synthetic metric set generated by ViT-Large.}
    \label{fig: vit-large}
\end{figure}

We present visualizations of synthetic metric sets produced by various ViT model variants, as shown from \cref{fig: vit-tiny} to \cref{fig: vit-large}. These visualizations clearly demonstrate that the synthetic images adeptly replicate the distribution found within the original dataset. Starting as mere Gaussian noise with randomly allocated class labels, these images undergo a remarkable transformation. They're gradually updated to display significant semantic richness, embodying object textures, contours, and intricate details. This progression emphatically underscores the proficiency of our image-generation methodology.

On the other hand, even though the images we generate are still far from being as realistic as actual samples, the compressed model's metric loss on synthetic data can still reflect its true performance on the test set very accurately. This balance between simplicity in generation and accuracy in performance assessment marks a significant stride in using synthetic data for model evaluation.

\begin{table}
    \centering
	\small
	\setlength{\tabcolsep}{2.5pt}
    \begin{tabular}{c|cccccc}
    \toprule
     & ViT-T & ViT-S & ViT-B & ViT-L & DeiT-B & Swin-B\\
    \midrule
    Latency (ms) & $296.2$ & $526.5$ & $985.5$ & $3204.2$ & $302.4$ & $390.1$\\
    \bottomrule
    \end{tabular}
\caption{The time of one iteration for different ViT variants to update synthetic images.}
\label{table: iter}
\end{table}

As shown in \cref{table: iter}, we also tested the latency of one iteration for different ViT variants to update synthetic images. The time is measured on a single RTX 3090 GPU. Compared to the training time, we can see that for different models, the total time taken to generate synthetic data is very short, amounting to less than 10\% of the duration of a single finetune after compressing one block.



\end{document}